\documentclass[lettersize,journal]{IEEEtran}
\usepackage{amsmath,amsfonts}
\usepackage{algorithmic}
\usepackage{algorithm}
\usepackage{array}
\usepackage[caption=false,font=normalsize,labelfont=sf,textfont=sf]{subfig}
\usepackage{textcomp}
\usepackage{stfloats}
\usepackage{url}
\usepackage{verbatim}
\usepackage{graphicx}
\usepackage{cite}
\usepackage{multirow}
\usepackage{listings}  
\usepackage{float}
\usepackage{multirow}
\usepackage{booktabs}

\usepackage[table]{xcolor}

\usepackage[citecolor=blue,linkcolor=blue]{hyperref}

\title{Complex Mixer for MedMNIST Classification Decathlon}

\author{\bf Zhuoran Zheng$^{1}$\quad Xiuyi Jia$^{1}$\thanks{correspondence to jiaxy@njust.edu.cn}\\[2mm]
$^{1}$CSE, Nanjing University of Science and Technology
}

\begin{document}

\maketitle

\begin{abstract}
With the development of the medical image field, researchers seek to develop a class of datasets to block the need for medical knowledge, such as \text{MedMNIST} (v2).
MedMNIST (v2) includes a large number of small-sized (28 $\times$ 28 or 28 $\times$ 28 $\times$ 28) medical samples and the corresponding expert annotations (class label).
The existing baseline model (Google AutoML Vision, ResNet-50+3D) can reach an average accuracy of over 70\% on MedMNIST (v2) datasets, which is comparable to the performance of expert decision-making.
Nevertheless, we note that there are two insurmountable obstacles to modeling on MedMNIST (v2):
1) the raw images are cropped to low scales may cause effective recognition information to be dropped and the classifier to have difficulty in tracing accurate decision boundaries; 
2) the labelers' subjective insight may cause many uncertainties in the label space.
To address these issues, we develop a Complex Mixer (C-Mixer) with a pre-training framework to alleviate the problem of insufficient information and uncertainty in the label space by introducing an incentive imaginary matrix and a self-supervised scheme with random masking.
Our method (incentive learning and self-supervised learning with masking) shows surprising potential on both the standard MedMNIST (v2) dataset, the customized weakly supervised datasets, and other image enhancement tasks.
\end{abstract}

\section{Introduction}
Currently, medical image analysis~\cite{anwar2018medical,cao2023swin,shen2017deep} based on machine learning can help healthcare professionals to make efficient decisions.
However, most of the medical image datasets require background knowledge from pattern recognition (image processing) and medical imaging, which is a great challenge for non-specialized researchers.

To clear this obstacle, \textit{MedMNIST} (v2)~\cite{yang2023medmnist} is proposed, which involves samples that can be deemed as a ``blind box'' for the researcher, where the pipeline (classifier) is built without priori knowledge.
The details of \textit{MedMNIST} (v2) are summarized and reported in Table~\ref{T1}.
We note that there are two key issues with \textit{MedMNIST} (v2) that may block the performance of the existing classifier.
1) \textit{MedMNIST} (v2) contains samples that are cropped from a medical image with high resolution to an image resolution of 28 $\times$ 28 (2D) or 28 $\times$ 28 $\times$ 28 (3D).
Predictably, this encoding approach (linear interpolation) loses a large number of local details, and the information about the lesion on the medical image is diluted.
To efficiently extract lesion-related pixels, several ViT-based classifiers are proposed~\cite{liu2022feature,manzari2023medvit}, which attempt to incorporate long-range modeling with some local learning algorithms to deeply mine each pixel relationship on lesion information.
However, the effectiveness of the modeling depends purely on the capacity of the network (MedViT~\cite{manzari2023medvit} with at least \textcolor{red}{10M} parameters), and the enormous number of parameters brought by ViT is difficult to deploy in real medical image analysis.
2) Since there is subjective insight in expert labeling, the 18 datasets (see Table~\ref{T1}) are inevitably mislabeled, which causes a large number of uncertainties in the label space~\cite{jiang2022deeply}.
Existing baseline algorithms do not model the uncertainty in the label space, which is one of the critical reasons for the limited performance of the classifier.
\begin{table*}[t] \scriptsize
	\begin{center}
		\vspace{-0mm}
		\caption{Data summary of \textit{MedMNIST} (v2) dataset. MC: Multi-Class. BC: Binary-Class. ML: Multi-Label. OR: Ordinal Regression.}
		\vspace{1mm}
		\label{T1}
		\resizebox{\linewidth}{!}{
			\begin{tabular}{l|c|c|c|l}
				\toprule
				\rowcolor{black!15}
				\textbf{Name}            & \textbf{Task(\#Classes/Labels)} & \textbf{\#Samples} & \textbf{\#Training/Validation/Test} & \multicolumn{1}{c}{\textbf{\#Resolution}}                                 \\
				\midrule
				\rowcolor{red!15}
				\multicolumn{5}{c}{\textbf{MedMNIST2D}}                                                                                                \\
				\midrule
				PathMNIST       & MC(9)                  & 107,180   & 89,996/10,004/7,180        & 3 $\times$ 224 $\times$ 224 $\rightarrow$ 3 $\times$ 28 $\times$ 28                 \\
				\midrule
				ChestMNIST      & ML(14) BC(2)           & 112,120   & 78,468/11,219/22,433       & 1 $\times$ 1024 $\times$ 1024 $\rightarrow$ 1 $\times$ 28 $\times$ 28               \\
				\midrule
				DermaMNIST      & MC(7)                  & 10,015    & 7,007/1,003/2,005          & 3 $\times$ 600 $\times$ 450 $\rightarrow$ 3 $\times$ 28 $\times$ 28                 \\
				\midrule
				OCTMNIST        & MC(4)                  & 109,309   & 97,477/10,832/1,000        & 1 $\times$ (384-1,536) $\times$ (277-512) $\rightarrow$ 1 $\times$ 28 $\times$ 28   \\
				\midrule
				PneumoniaMNIST  & BC(2)                  & 5,856     & 4,708/524/624              & 1 $\times$ (384-2,916) $\times$ (127-2,713) $\rightarrow$ 1 $\times$ 28 $\times$ 28 \\
				\midrule
				RetinaMNIST     & OR(5)                  & 1,600     & 1,080/120/400              & 3 $\times$ 1,736 $\times$ 1,824 $\rightarrow$ 3 $\times$ 28 $\times$ 28             \\
				\midrule
				BreastMNIST     & BC(2)                  & 780       & 546/78/156                 & 1 $\times$ 500 $\times$ 500 $\rightarrow$ 1 $\times$ 28 $\times$ 28                 \\
				\midrule
				BloodMNIST      & MC(8)                  & 17,092    & 11,959/1,712/3,421         & 3 $\times$ 200 $\times$ 200 $\rightarrow$ 3 $\times$ 28 $\times$ 28                 \\
				\midrule
				TissueMNIST     & MC(8)                  & 236,386   & 165,466/23,640/47,280      & 32 $\times$ 32 $\times$ 7  $\rightarrow$ 1 $\times$ 28 $\times$ 28                   \\
				\midrule
				OrganAMNIST     & MC(11)                 & 58,850    & 34,581/6,491/17,788        & - $\rightarrow$ 1 $\times$ 28 $\times$ 28                             \\
				\midrule
				OrangCMNIST     & MC(11)                 & 23,660    & 13,000/2,392/8,268         & - $\rightarrow$ 1 $\times$ 28 $\times$ 28                             \\
				\midrule
				OrganSMNIST     & MC(11)                 & 25,221    & 13,940/2,452/8,829         & - $\rightarrow$ 1 $\times$ 28 $\times$ 28                             \\
				\midrule
				\rowcolor{red!15}
				\multicolumn{5}{c}{\textbf{MedMNIST3D}}                                                                                                \\
				\midrule
				OrganMNIST3D    & MC(11)                 & 1,743     & 972/161/610                & - $\rightarrow$ 28 $\times$ 28 $\times$ 28                            \\
				\midrule
				NoduleMNIST3D   & BC(2)                  & 1,633     & 1,158/165/310              & 1MM $\times$ 1MM $\times$ 1MM $\rightarrow$ 28 $\times$ 28 $\times$ 28              \\
				\midrule
				AdrenalMNIST3D  & BC(2)                  & 1,584     & 1,188/98/298               & 64M $\times$ 64M $\times$ 64M $\rightarrow$ 28 $\times$ 28 $\times$ 28              \\
				\midrule
				FractureMNIST3D & MC(3)                  & 1,370     & 1,207/103/240              & 64M $\times$ 64M $\times$ 64M $\rightarrow$ 28 $\times$ 28 $\times$ 28              \\
				\midrule
				VesselMNIST3D   & BC(2)                  & 1,909     & 1,335/192/382              & - $\rightarrow$ 28 $\times$ 28 $\times$ 28                            \\
				\midrule
				SynapseMNIST3D  & BC(2)                  & 1,759     & 1,230/177/352              & 1024 $\times$ 1024 $\times$ 1024 $\rightarrow$ 28 $\times$ 28 $\times$ 28  \\  
				\bottomrule
				
			\end{tabular}
		}
		\vspace{-4mm}
	\end{center}
\end{table*}

Considering together, we propose the \textbf{incentive learning} (a novel machine learning paradigm) and a self-supervised pre-training method with random masking.
The key to our approach is to learn incentivized imaginary information, which is then enforced on the raw image to generate a complex space.
Finally, Complex Mixer (C-Mixer) and Pearson curves are used for constructing the learner.
We develop this algorithm from two insights:
i) Incentive learning: some conditional noise (the mean and variance of Gaussian noise are learned parameters) is added to the input information that may reduce the learner's modeling cost and complement some recognition-friendly signals~\cite{li2022positive}.
ii) High-dimensional space learning: the imaginary numbers implicitly involve properties such as rotation and phase change, and combining them with the real number space can generate a complex space to learn a robust decision plane for the classifier in higher dimensions~\cite{choi2019phase,HuLLXZFWZX20,trabelsi2017deep}.
In addition, we design a pre-training method with random masking that can enforce a latent regularization term on the C-Mixer so that it has consistently determined for similar samples without the disturbance of uncertainty in the label space.
Extensive experimental results demonstrate that our method outperforms existing state-of-the-art methods on standard \textit{MedMNIST} (v2) as well as on custom semi-supervised datasets by an average of 5-9\%.
Besides, we also try to provide the existing algorithm (image enhancement task) with an incentivized Gaussian noise or a pre-training method with random masking, and the experiments show that both of them have different degrees of gain.
The contributions of this paper are summarized as follows:
\begin{itemize}
	\vspace{-1mm}
	\item We propose the incentive learning as well as a self-supervised method with random masking, which can overcome two key problems that exist in \textit{MedMNIST} (v2).
	%
	%
	\vspace{-1mm}
	\item We introduce the complex field (input an incentivized signal), which to our knowledge is the first attempt to model a learner by boosting the number field dimension in a small-scale sample.
	\vspace{-1mm}
	\item Experimental results on \textit{MedMNIST} (v2) demonstrate the proposed algorithm performs favorably against the state-of-the-art image recognition  methods. In addition, our proposed learning paradigm has significant gains in enhancement tasks.
\end{itemize}

\begin{figure*}[t]
	\centering
	\includegraphics[width=0.920\textwidth]{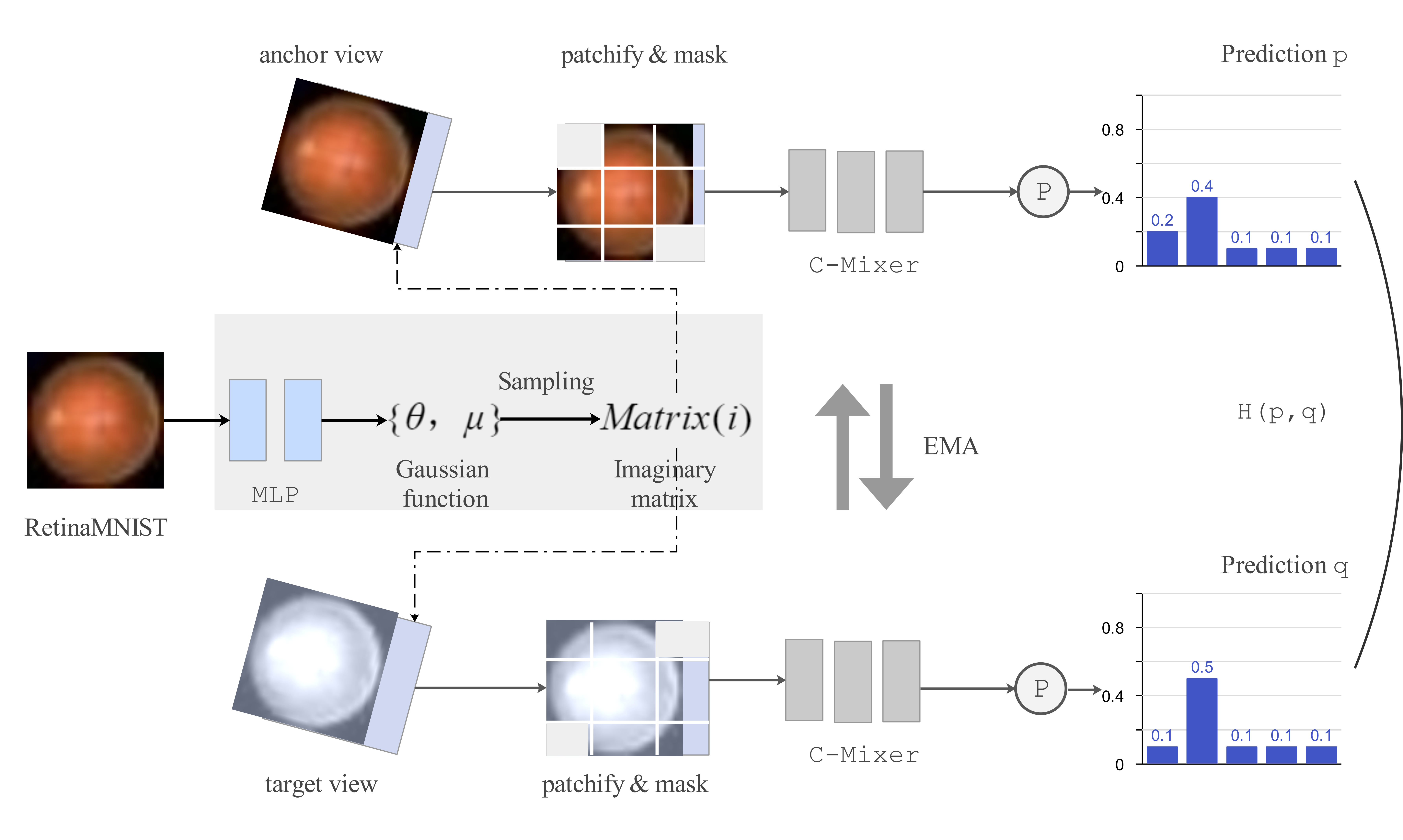}
	\vspace{-2mm}
	\caption{
		This figure shows the architecture of the proposed self-supervised learning framework with random masking.
		The anchor view and target view are randomly masked with a certain probability of being input into the C-Mixer. \texttt{P} aims at projecting the complex number domain into the real number domain, and EMA represents parameter sharing.
	}
	\vspace{-2mm}
	\label{frameworks}
\end{figure*}

\section{Method}
Our approach involves three key components, incentive learning, a C-Mixer network, and a self-supervised learning framework.
Incentive learning is intended to tackle the problem of insufficient input information by introducing a conditional noise; C-Mixer is a learner used to extract features from the complex space; a self-supervised learning framework is employed as a latent regularization strategy to overcome the problem of uncertainty in the label space.

\noindent \textbf{Incentive learning.}		
Li~\cite{li2022positive} proposed and confirmed that customized conditional noise can reduce the learning load of the learner and even counteract the negative noise.
Inspired by this, we propose a new machine-learning paradigm termed incentive learning. 
This framework can be formalized as:
\begin{equation}
	\label{e1}
	y = f(\underbrace{\textbf{x} \quad \Delta \quad \textbf{x}^{*}}_{\text{\textcolor{blue}{Information fusion}}}), 
\end{equation}
where \textbf{x} denotes the raw sample, $\textbf{x}^{*}$ denotes the customized noise from a distribution $\mathbb{D}$, and $f$ denotes the leaner.
The parameters of distribution $\mathbb{D}$ comes from a learner ($\texttt{MLPs}$ or $\texttt{CNNs}$) conducting modeling on $\textbf{x}$ to obtain the parameters $\Theta$ (for the Gaussian distribution, the parameters are $\mu$ and $\sigma$) of the prior distribution $\mathbb{D}$.
In addition, $\Delta$ in Equation~\ref{e1} denotes an operator for information fusion, such as $+$ or feature map stacking.

In this paper, we assume that the noise with condition $\textbf{x}^{*}$ comes from a Gaussian distribution $\mathbb{D}_G$ and that it is an imaginary space $\Omega(i)$.
This is because images that are drastically downsampled are hard for the learner to mine for distinguishing features only in the \textbf{RGB domain}.
The imaginary space $\Omega(i)$ provides a solution strategy from another view because it provides information like frequency domain variations as well as rotations, phases, and other variations.
Specifically, we start by using the \texttt{Flatten} operator to unfold a vector on the input image $I$, next extract the deep semantics of the vector by employing an \texttt{MLP}, and then generate two parameters ($\mu$ and $\sigma$) by a normalization operator.
Finally, a matrix or vector $\textbf{x}^{*}$ with imaginary numbers $i$ is sampled by knowing the two parameters ($\mu$ and $\sigma$) of the Gaussian function $\mathbb{D}_G$.
Here, the dimensions of the generated matrix $\textbf{x}^{*}$ are the same as the raw image $I$.	
This procedure can be formalized as follows:
\begin{equation}
	\textbf{x}^{*}(i) \sim \mathbb{D}_G(\underbrace{\texttt{Tanh}(\texttt{MLP}(\texttt{Flatten}(I)))}_{\text{\textcolor{blue}{Parameter generation}}}),
\end{equation}
where the role of $\texttt{Tanh}$ is normalized.	
Note that $\texttt{MLP}$ can also be replaced by $\texttt{CNNs}$ with global pooling.	

So far, $\Delta$ requires thoughtful design and deliberation to form a high-quality input signal.
Here, we use $\textbf{x}^{*}(i)$ as a channel of the input signal with the help of the $\texttt{cat}$ operator.	
After that, the real and imaginary matrices are fused by using the $\texttt{C-Mixer}$.

\noindent \textbf{C-Mixer.}	
How to build an efficient $f$?
Up to now, the existing state-of-the-art models consider the long-range dependence of exploiting pixels, and here we use MLP-Mixer~\cite{TolstikhinHKBZU21} as a baseline to develop the pipeline.
A standard MLP-Mixer starts acting on $\text{\textbf{X}} \in \mathbb{R}^{S \times C}$ ($S$ denotes the sequence of inputs and $C$ denotes the dimension of the hidden layer) by using a split-patch scheme.
All patches are linearly projected with the same projection matrix $\text{\textbf{W}}$. 
MLP-Mixer consists of multiple layers of identical size, and each layer consists of two MLP blocks.
Each MLP block contains two \texttt{FC} layers and a \texttt{GELU} ($\sigma$) conducted independently to each row of its input data tensor.
Mixer layers can be written as follows:
\begin{equation}
	\begin{aligned}
		\text{\textbf{U}}_{\ast,i} = \text{\textbf{X}} + \text{\textbf{W}}_{2} \sigma(\text{\textbf{W}}_{1}\text{LayerNorm}(\text{\textbf{X}}_{\ast,i})), \quad \text{for i} = \text{1} \text{...} C,
		\\
		\text{\textbf{Y}}_{\ast,j} = \text{\textbf{U}} + \text{\textbf{W}}_{4} \sigma(\text{\textbf{W}}_{3}\text{LayerNorm}(\text{\textbf{U}}_{\ast,j})), \quad \text{for j} = \text{1} \text{...} S.
	\end{aligned}
\end{equation}
Since our network is fed with a tensor ($\text{\textbf{h}} = \text{\textbf{a}} + \text{\textbf{b}}i $) in the form of a complex number, a modification of $\text{\textbf{W}}$ is required.
Here, $\text{\textbf{W}}$ can be designed to contain two affine matrices ($\text{\textbf{W}}$ = $\text{\textbf{A}}$ $+$ $\text{\textbf{B}}i$).
The process of affine transformation can be written in the following form:
\begin{equation}
	\text{\textbf{W}}\text{\textbf{h}} = (\text{\textbf{Aa}} - \text{\textbf{Bb}}) + i(\text{\textbf{Ba}} + \text{\textbf{Ab}}).
\end{equation}
In addition, for Mixer, the activation function $\sigma$ is replaced by $\mathbb{C}\text{ReLU}$.
\begin{equation}
	\mathbb{C}\text{ReLU} = \text{ReLU}(a) + i\text{ReLU}(b).
\end{equation}
In general, the affine transformation $\text{\textbf{W}}$ and the activation function $\sigma$ are customized privately based on MLP-Mixer, called \texttt{C-Mixer}.
Note that the labels in \textit{MedMNIST} (v2) are all integer values $\mathbb{Z}$, and the network is constructed on the complex domain, so the features require to be transformed to the real field $\mathbb{R}$ to be used for inference.
To address this problem, we employ generalized Pearson curves \texttt{p} to achieve the conversion between number fields.
This operator can be formalized as follows:
\begin{equation}
	\texttt{p} = \texttt{Tanh}(\text{real}(y) + \text{imag.}(y)),
\end{equation}
where $\text{real}$ and $\text{imag.}$ denote the real and imaginary numbers that extract the complex number field $y$, respectively.
C-Mixer has the ability to model long ranges and fewer FLOPs than existing complex deep networks.

\noindent \textbf{Self-supervised learning framework.}		
In this paper, there is a significant issue that still requires addressing i.e., uncertainty in the label space.	
An obvious solution to the idea is that similar inputs should have similar representations.
From this, we develop a self-supervised learning method, which aims to keep the deep semantics learned by the network consistent for similar images.
As shown in Figure~\ref{frameworks}, the anchor view is the input image of the model, the target view is the augmentation at the base of the anchor view, the parameters of \texttt{C-Mixer} are shared, and \texttt{P} denotes the generalized Pearson curve.
There is a key innovation in the overall architecture, where the anchor view and target view are randomly masked by some pixels of the image before being input into \texttt{C-Mixer}.
Self-supervised learning framework can be written as follows:	
\begin{equation}
	\begin{aligned}
		\text{min}\texttt{H(p,q)} = \texttt{H}(\texttt{P}(\texttt{C-Mixer}(\text{Mask}(\text{anchor view}))),
		\\
		\texttt{P}(\texttt{C-Mixer}(\text{Mask}(\text{target view})))),
	\end{aligned}
\end{equation}	
where \texttt{H} denotes cross-entropy, Mask is a global random masking method rather than a locally learned operator.

\noindent \textbf{Loss function.}		
Since there are mainly two different tasks in \textit{MedMNIST} (v2), the loss function $\mathcal{L}$ uses in these two tasks (\{multi-label, binary class\}, \{binary/multi--lass and ordinal regression\}) is different as well. 
1) For multi-label, binary class, we employ BCEWithLogitsLoss.
\begin{equation}
	\begin{aligned}
		\text{L}_{\text{i}} = -[y_{\text{i}} \times log(\hat{y_{\text{i}}}) + (1-y_{\text{i}}) \times log(1-\hat{y}_{\text{i}})],
		\\
		\mathcal{L} = \{\text{L}_{\text{1}}, ..., 	\text{L}_{\text{n}}\}.
	\end{aligned}	
\end{equation}

2) For binary/multi-class and ordinal regression, we employ CrossEntropyLoss.
\begin{equation}
	\mathcal{L} = -\frac{1}{n}\sum_{\text{i}=0}^{n}[y_{\text{i}} \times log(\hat{y_{\text{i}}}) + (1-y_{\text{i}}) \times log(1-\hat{y}_{\text{i}})].
\end{equation}

\section{Experiments}
This section involves an introduction to the dataset, evaluation methods, quantitative analysis, ablation, and discussion and analysis.

\noindent \textbf{Dataset.}
The \textit{MedMNIST} (v2) dataset includes 12 2D and 6 3D standardized datasets from carefully selected sources covering primary data modalities (X-ray, OCT, ultrasound, CT, and electron microscope), diverse classification tasks (binary/multi-class, ordinal regression, and multi-label) and dataset scales (from 100 to 100,000).
We illustrate the landscape of \textit{MedMNIST} (v2) in Figure~\ref{T1}.
The data files (each subset is saved in \textit{NumPy} (npz) format) of \textit{MedMNIST} (v2) dataset can be accessed at \url{https://medmnist.com/}.

\noindent \textbf{Evaluation metrics.}
Following the existing work, we select AUC and ACC as evaluation methods.
\textbf{ACC}~\cite{zhou2021machine}: (TP + TN)/(TP + FP + FN + TN); TP: True Positive (the predicted values of the sample match the true values and are all positive), FP: False Positive (the sample prediction is positive and the true value is negative), FN: False Negative (the sample predicts a negative value while the true value is positive), TN: True Negative (the predicted values of the sample match the true values and are all negative).
\textbf{AUC}~\cite{zhou2021machine}: The AUC value is equivalent to the probability that a randomly chosen positive example is ranked higher than a randomly chosen negative example. 
Usually, the AUC is computed (area under ROC) before a ROC curve is prepared.

\begin{table*}[!htb] \tiny
	\caption{{\bf Overall performance of \textit{MedMNIST} (v2) \textcolor{red}{[Fully-supervised learning]}} in metrics of AUC and ACC, using ResNet-18 / ResNet-50 with resolution $28$ and $224$, auto-sklearn, AutoKeras, Google AutoML Vision, FPVT, MedVIT, and Ours.}
	\label{tab:2dResults}
	\vspace{-1mm}
	\begin{center}
		\resizebox{\linewidth}{!}{
			\begin{tabular}{@{}ccccccccccccc@{}}
				\toprule
				\multirow{2}{*}{Methods} &
				\multicolumn{2}{c}{PathMNIST} &
				\multicolumn{2}{c}{ChestMNIST} &
				\multicolumn{2}{c}{DermaMNIST} &
				\multicolumn{2}{c}{OCTMNIST} &
				\multicolumn{2}{c}{PneumoniaMNIST} &
				\multicolumn{2}{c}{BloodMNIST} 
				\\
				& AUC & ACC & AUC & ACC & AUC & ACC & AUC & ACC & AUC & ACC & AUC & ACC\\ \midrule
				ResNet-18 (28)     & 0.972 & 0.844 & 0.706 & 0.947 & 0.899 & 0.721 & 0.951 & 0.758 & 0.957 & 0.843 & 0.998 & 0.958\\
				ResNet-18 (224)        & 0.978 & 0.860 & 0.713 & 0.948 & 0.896 & 0.727 & 0.960 & 0.752 & 0.970 & 0.861  &0.998 &0.958\\
				ResNet-50 (28)        & 0.979 & 0.864 & 0.692 & 0.947 & 0.886 & 0.710 & 0.939 & 0.745 & 0.949 & 0.857 & 0.997 & 0.956\\
				ResNet-50 (224)       & 0.978 & 0.848 & 0.706 & 0.947 & 0.895 & 0.719 & 0.951 & 0.750 & 0.968 & 0.896 & 0.997 & 0.950 \\
				auto-sklearn         & 0.500 & 0.186 & 0.647 & 0.642 & 0.906 & 0.734 & 0.883 & 0.595 & 0.947 & 0.865 & 0.984 & 0.878\\
				AutoKeras           & 0.979 & 0.864 & 0.715 & 0.939 & 0.921 & 0.756 & 0.956 & 0.736 & 0.970 & 0.918 & 0.998 & 0.961\\
				Google AutoML Vision  & 0.982 & 0.811 & 0.718 & 0.947 & 0.925 & 0.766 & 0.965 & 0.732 & 0.993 & 0.941  & 0.998 & 0.966\\
				
				FPVT  & 0.994& 0.918 & 0.725 & 0.948 & 0.923 & 0.766 & 0.968 & 0.813 & 0.973 & 0.896 & 0.975 & 0.944\\
				MedVIT-T  & 0.994& 0.938 & 0.786 & 0.956 & 0.914 & 0.768 & 0.961 & 0.767 & 0.993 & 0.949 & 0.996 & 0.950\\
				MedVIT-S  & 0.993& 0.942 & 0.791 & 0.954 & 0.937 & 0.780 & 0.960 & 0.782 & 0.995 & 0.961 & 0.997 & 0.951\\
				MedVIT-L  & 0.984& 0.933 & 0.805 & 0.959 & 0.920 & 0.773 & 0.945 & 0.761 & 0.991 & 0.921 & 0.996 & 0.954\\
				Ours  & \bf0.995& \bf0.943 & \bf0.822 & \bf0.963 & \bf0.940 & \bf0.833 & \bf0.969 & \bf0.840 & \bf0.997 & \bf0.966 & \bf0.998 & \bf0.979\\
				
				\midrule
				\multirow{2}{*}{Methods} &
				\multicolumn{2}{c}{RetinaMNIST} &
				\multicolumn{2}{c}{BreastMNIST} &
				\multicolumn{2}{c}{OrganMNIST\_A} &
				\multicolumn{2}{c}{OrganMNIST\_C} &
				\multicolumn{2}{c}{OrganMNIST\_S} &
				\multicolumn{2}{c}{TissueMNIST}   \\
				& AUC & ACC & AUC & ACC & AUC & ACC & AUC & ACC & AUC & ACC & AUC & ACC\\ \midrule
				ResNet-18 (28)         & 0.727 & 0.515 & 0.897 & 0.859 & 0.995 & 0.921 & 0.990 & 0.889 & 0.967 & 0.762 & 0.930 & 0.676\\
				ResNet-18 (224)       & 0.721 & 0.543 & 0.915 & 0.878 & 0.997 & 0.931 & 0.991 & 0.907 & 0.974 & 0.777 & 0.933 & 0.681\\
				ResNet-50 (28)         & 0.719 & 0.490 & 0.879 & 0.853 & 0.995 & 0.916 & 0.990 & 0.893 & 0.968 & 0.746 & 0.931 & 0.680\\
				ResNet-50 (224)        & 0.717 & 0.555 & 0.863 & 0.833 & 0.997 & 0.931 & 0.992 & 0.898 & 0.970 & 0.770 &0.932 & 0.680\\
				auto-sklearn        & 0.694 & 0.525 & 0.848 & 0.808 & 0.797 & 0.563 & 0.898 & 0.676 & 0.855 & 0.601 & 0.828 & 0.532\\
				AutoKeras           & 0.655 & 0.420 & 0.833 & 0.801 & 0.996 & 0.929 & 0.992 & 0.915 & 0.972 & 0.803 & 0.941 & 0.703\\
				Google AutoML Vision  & 0.762 & 0.530    &  0.932 & 0.865    & 0.988 & 0.818 & 0.986 & 0.861 & 0.964 & 0.706  &0.924 & 0.673\\
				FPVT                  & 0.753    & 0.568 & 0.938     & 0.891 & 0.997 & 0.935 & 0.993 & 0.903 & 0.976 & 0.785  &0.923 & 0.717\\
				MedVIT-T         & 0.752    & 0.534 & 0.934     & 0.896 & 0.995 & 0.931 & 0.991 & 0.901 & 0.972 & 0.789  &0.943 & 0.703\\
				MedVIT-S         & 0.773    & 0.561 & 0.938     & 0.897 & 0.996 & 0.928 & 0.993 & 0.916 & 0.987 & 0.805  &\bf0.952 & 0.731\\
				MedVIT-L         & 0.754    & 0.552 & 0.929     & 0.883 & 0.997 & 0.943 & \bf0.994 & \bf0.922 & 0.973 & 0.806  &0.935 & 0.699\\
				Ours             & \bf0.777    & \bf0.570 & \bf0.939     & \bf0.899 & \bf0.997 & \bf0.951 & 0.993 & 0.917 & \bf0.977 & \bf0.810  &0.944 &\bf0.739\\
				\midrule
				\multirow{2}{*}{Methods} &
				\multicolumn{2}{c}{OrganMNIST3D} &
				\multicolumn{2}{c}{NoduleMNIST3D} &
				\multicolumn{2}{c}{FractureMNIST3D} &
				\multicolumn{2}{c}{AdrenalMNIST3D} &
				\multicolumn{2}{c}{VesselMNIST3D} &
				\multicolumn{2}{c}{SynapseMNIST3D} 
				\\
				& AUC & ACC & AUC & ACC & AUC & ACC & AUC & ACC & AUC & ACC & AUC & ACC\\ \midrule
				ResNet-18 + 2.5D     & 0.977 & 0.788 & 0.838 & 0.835 & 0.587 & 0.451 & 0.718 & 0.772 & 0.748 & 0.846 &0.634 &0.696  \\
				ResNet-18 + 3D        & \bf0.996 & 0.907 & 0.863 & 0.844 & 0.712 & 0.508 & 0.827 & 0.721 & 0.874 & 0.877 &0.820 &0.745  \\
				ResNet-18 + ACS        & 0.994 & 0.900 & 0.873 & 0.847 & 0.714 & 0.497 & 0.839 & 0.754 & 0.930 & 0.928 &0.705 &0.722   \\
				ResNet-50 + 2.5D       & 0.974 & 0.769 & 0.835 & 0.848 & 0.552 & 0.397 & 0.732 & 0.763 & 0.751 & 0.877 &0.669 &0.735  \\
				ResNet-50 + 3D       & 0.994 & 0.883 & 0.875 & 0.847 & 0.725 & 0.494 & 0.828 & 0.745 & 0.907 & 0.918 &0.851 &0.795  \\
				ResNet-50 + ACS      & 0.994 & 0.889 & 0.886 & 0.841 & \bf0.750 & 0.517 & 0.828 & 0.758 & 0.912 & 0.858 &0.719 & 0.709  \\
				auto-sklearn         & 0.977 & 0.814 & 0.914 & 0.874 & 0.628 & 0.453 & 0.828 & \bf0.802 & 0.910 & 0.915 &0.631 &0.730  \\
				AutoKeras           & 0.979 & 0.804 & 0.844 & 0.834 & 0.642 & 0.458 & 0.804 & 0.705 & 0.773 & 0.894 &0.538 &0.724 \\
				FPVT           & 0.923 & 0.800 & 0.814 & 0.822 & 0.640 & 0.438 & 0.801 & 0.704 & 0.770 & 0.888 &0.530 &0.712 \\
				MedViT (T/S/L)           & - & - & - & - & - & - & - & - & - & - &- &- \\
				Ours  & 0.995& \bf0.912 & \bf0.915 & \bf0.860 & 0.729 & \bf0.660 & \bf0.969 & 0.801 & \bf0.932 & \bf0.940 & \bf0.866 &\bf0.820\\
				
				\bottomrule
			\end{tabular}
		}
		\vspace{-4mm}
	\end{center}
\end{table*}

\noindent \textbf{Comparison.}
We conduct some comparison methods on fully-supervised, semi-supervised, and weakly-supervised tasks.
Comparison method: 
For \textit{MedMNIST2D} (v2), \textcolor{blue}{ResNets} (including 4 residual layers, batch normalization, and \texttt{ReLU} activation) with a simple early-stopping strategy on the validation are set as baseline methods.
We use cross-entropy and set the batch size to 128 during the model training.
We utilize an AdamW optimizer~\cite{loshchilov2017fixing} with an initial learning rate of 0.001 and train the network for 100 epochs, delaying the learning rate by 0.1 after 50 and 75 epochs.
\textcolor{blue}{FPVT}~\cite{liu2022feature} uses the Cutout to alleviate the problem of over-fitting. In addition, FPVT uses the AdamW optimizer, and the learning rate is set to $1 \times 10 ^{-3}$. For a fair comparison, here we also use the early-stopping scheme.
%
Since \textcolor{blue}{MedViT}~\cite{manzari2023medvit} does not release the code in the public platform (Github), we excerpt the reported results in our paper.
In addition, MedViT also does not release the test report in the 3D dataset.
For \textit{MedMNIST3D} (v2), we conduct \textcolor{blue}{ResNet-18/ResNet-50} with 2.5D/3D/ACS with a simple early-stopping scheme on the validation set as baseline methods.
The rest of the training configuration for \textcolor{blue}{ResNet-18/ResNet-50} is consistent with the 2D image processing case, and the same case for \textcolor{blue}{FPVT} and \textcolor{blue}{MedViT}.
In addition, we run auto-sklearn and AutoKeras on both \textit{MedMNIST2D} and \textit{MedMNIST3D}, and Google AutoML Vision on \textit{MedMNIST2D} only.
All comparison methods are conducted on the RTX3090 GPU shader and the experimental platform is PyTorch 1.2 (Python 3.8).
Experimental tasks: We set up three tasks, fully supervised, semi-supervised, and weakly-supervised (label space with error label).
The fully-supervised task follows exactly the divisions of the dataset setting in Table~\ref{T1}.
The semi-supervised task uses only 10\% of the training data with labels in Table~\ref{T1}, and the rest of the training samples are added to the test set.
The weakly-supervised task is to assign a certain number of error labels to the label space in the 10\% of the training set.
To be fair, FPVT and the ResNets run our proposed self-supervised training scheme on semi-supervised and weakly-supervised tasks.
FPVT and ResNets perform pre-training using the same optimizer, number of training steps, and learning rate as our method.
Table~\ref{tab:2dResults}-\ref{tab:4dResults} show the overall performance of the above models.
\begin{table*}[!htb] \tiny
	\caption{{\bf Overall performance of \textit{MedMNIST} (v2) \textcolor{red}{[Semi-supervised learning]}} in metrics of AUC and ACC, using ResNet-18 / ResNet-50 with resolution $28$ and $224$, auto-sklearn, AutoKeras, Google AutoML Vision, FPVT and Ours.}
	\label{tab:3dResults}
	\vspace{-1mm}
	\begin{center}
		\resizebox{\linewidth}{!}{
			\begin{tabular}{@{}ccccccccccccc@{}}
				\toprule
				\multirow{2}{*}{Methods} &
				\multicolumn{2}{c}{PathMNIST} &
				\multicolumn{2}{c}{ChestMNIST} &
				\multicolumn{2}{c}{DermaMNIST} &
				\multicolumn{2}{c}{OCTMNIST} &
				\multicolumn{2}{c}{PneumoniaMNIST} &
				\multicolumn{2}{c}{BloodMNIST} 
				\\
				& AUC & ACC & AUC & ACC & AUC & ACC & AUC & ACC & AUC & ACC & AUC & ACC\\ \midrule
				ResNet-18 (28)     & 0.673 & 0.522 & 0.616 & 0.643 & 0.709 & 0.539 & 0.641 & 0.728 & 0.737 & 0.533 & 0.540 & 0.608\\
				ResNet-18 (224)        & 0.630 & 0.501 & 0.638 & 0.649 & 0.511 & 0.532 & 0.644 & \bf0.794 & 0.634 & 0.509 & 0.428 & 0.302\\
				ResNet-50 (28)        & 0.399 & 0.411 & 0.426 & 0.575 & 0.677 & 0.536 & 0.467 & 0.541 & 0.637 & 0.500 & 0.522 & 0.608\\
				ResNet-50 (224)       & 0.333 & 0.522 & 0.616 & 0.640 & 0.709 & 0.499 & 0.297 & 0.598 & 0.563 & 0.337 & 0.524 & 0.608\\
				auto-sklearn         & 0.419 & 0.555 & 0.432 & 0.613 & 0.712 & 0.533 & 0.641 & 0.728 & 0.737 & 0.532 & 0.548 & 0.517\\
				AutoKeras           & 0.390 & 0.623 & 0.655 & 0.546 & 0.564 & 0.521 & 0.600 & 0.700 & 0.714 & 0.488 & 0.521 & 0.625\\
				Google AutoML Vision  & 0.477 & 0.499 & 0.510 & 0.511 & 0.399 & 0.300 & 0.246 & 0.533 & 0.637 & 0.530 & 0.622 & 0.678\\
				
				FPVT   & 0.699 & 0.539 & \bf0.689 & 0.687 & 0.641& 0.588 & 0.659 & 0.660 & 0.788 & 0.435 & 0.441 & 0.638\\
				Ours  & \bf0.701 & \bf0.644 & 0.630 & \bf0.688 & \bf0.711 & \bf0.621 & \bf0.699 & 0.728 & \bf0.799 & \bf0.566 & \bf0.598 & \bf0.700\\
				
				\bottomrule
			\end{tabular}
		}
		\vspace{-3mm}
	\end{center}
\end{table*}

\begin{table*}[!htb] \tiny
	\caption{{\bf Overall performance of \textit{MedMNIST} (v2) \textcolor{red}{[Weakly-supervised learning]}} in metrics of AUC and ACC, using ResNet-18 / ResNet-50 with resolution $28$ and $224$, auto-sklearn, AutoKeras, Google AutoML Vision, FPVT, MedVIT, and Ours.}
	\label{tab:4dResults}
	\vspace{-1mm}
	\begin{center}
		\resizebox{\linewidth}{!}{
			\begin{tabular}{@{}ccccccccccccc@{}}
				\toprule
				\multirow{2}{*}{Methods} &
				\multicolumn{2}{c}{PathMNIST} &
				\multicolumn{2}{c}{ChestMNIST} &
				\multicolumn{2}{c}{DermaMNIST} &
				\multicolumn{2}{c}{OCTMNIST} &
				\multicolumn{2}{c}{PneumoniaMNIST} &
				\multicolumn{2}{c}{BloodMNIST} 
				\\
				& AUC & ACC & AUC & ACC & AUC & ACC & AUC & ACC & AUC & ACC & AUC & ACC\\ \midrule
				ResNet-18 (28)     & 0.970 & 0.823 & 0.700 & 0.941 & 0.846 & 0.711 & 0.950 & 0.730 & 0.953 & 0.840 & 0.990 & 0.932\\
				ResNet-18 (224)    & 0.971 & 0.860 & 0.702 & 0.943 & 0.890 & 0.721 & 0.952 & 0.753 & 0.960 & 0.848  &0.990 &0.955\\
				ResNet-50 (28)     & 0.971 & 0.833 & 0.691 & 0.942 & 0.883 & 0.705 & 0.923 & 0.744 & 0.941 & 0.833 & 0.989 & 0.950\\
				ResNet-50 (224)    & 0.973 & 0.841 & 0.676 & 0.929 & 0.890 & 0.713 & 0.944 & 0.702 & 0.960 & 0.893 & 0.972 & 0.935 \\
				auto-sklearn       & 0.444 & 0.386 & 0.640 & 0.625 & 0.886 & 0.730 & 0.843 & 0.591 & 0.940 & 0.863 & 0.982 & 0.870\\
				AutoKeras          & 0.951 & 0.860 & 0.711 & 0.932 & 0.910 & 0.755 & 0.950 & 0.731 & 0.965 & 0.911 & 0.994 & 0.950\\
				Google AutoML Vision  & 0.981 & 0.833 & 0.710 & 0.941 & 0.920 & 0.749 & 0.932 & 0.722 & 0.990 & 0.930  & 0.992 & 0.957\\
				
				FPVT  & 0.965& 0.900 & 0.715 & 0.940 & 0.911 & 0.753 & 0.952 & 0.769 & 0.930 & 0.892 & 0.970 & 0.942\\
				Ours  & \bf0.992& \bf0.939 & \bf0.816 & \bf0.959 & \bf0.931 & \bf0.826 & \bf0.966 & \bf0.835 & \bf0.988 & \bf0.960 & \bf0.992 & \bf0.989\\
				
				\bottomrule
			\end{tabular}
		}
		\vspace{-3mm}
	\end{center}
\end{table*}

\begin{table*}[!htb] \tiny
	\caption{{\bf Overall performance of \textit{MedMNIST} (v2) \textcolor{red}{[Ablation experiments]}} in metrics of AUC and ACC.}
	\label{tab:5Results}
	\vspace{-1mm}
	\begin{center}
		\resizebox{\linewidth}{!}{
			\begin{tabular}{@{}ccccccccccccc@{}}
				\toprule
				\multirow{2}{*}{Methods} &
				\multicolumn{2}{c}{PathMNIST} &
				\multicolumn{2}{c}{ChestMNIST} &
				\multicolumn{2}{c}{DermaMNIST} &
				\multicolumn{2}{c}{OCTMNIST} &
				\multicolumn{2}{c}{PneumoniaMNIST} &
				\multicolumn{2}{c}{BloodMNIST} 
				\\
				& AUC & ACC & AUC & ACC & AUC & ACC & AUC & ACC & AUC & ACC & AUC & ACC\\ \midrule
				w/o ssl & 0.990& 0.941 & 0.819 & 0.952 & 0.938 & 0.831 & 0.969 & 0.832 & 0.990 & 0.962 & 0.996 & 0.970\\
				w/o rm  & 0.991& 0.942 & 0.820 & 0.959 & 0.936 & 0.831 & 0.962 & 0.835 & 0.991 & 0.964 & 0.996 & 0.977\\
				w/o il  & 0.983& 0.930 & 0.811 & 0.962 & 0.931 & 0.826 & 0.961 & 0.830 & 0.989 & 0.962 & 0.988 & 0.967\\
				w/o \texttt{p}(r) & 0.994& 0.942 & 0.820 & 0.961 & 0.938 & 0.830 & 0.962 & 0.835 & 0.995 & 0.962 & 0.992 & 0.978\\
				w/o \texttt{p}(i) & 0.991 & 0.942 & 0.822 & 0.960 & 0.942 & 0.830 & 0.962 & 0.837 & 0.992 & 0.965 & 0.992 & 0.973\\
				Ours  & \bf0.995& \bf0.943 & \bf0.822 & \bf0.963 & \bf0.940 & \bf0.833 & \bf0.969 & \bf0.840 & \bf0.997 & \bf0.966 & \bf0.998 & \bf0.979\\
				
				\bottomrule
			\end{tabular}
		}
		\vspace{-6mm}
	\end{center}
\end{table*}

\noindent \textbf{Implementation details.}
1) Specifications of the $\texttt{C-Mixer}$ architectures.
We follow the layout of a standard MLP-Mixer with some key parameters, Number of layers: 8, Patch resolution $\text{P} \times \text{P}$: 8 $\times$ 8, Hidden size $C$: 218, Sequence length $S$: 49, Parameters (M): 6.
Each layer in $\texttt{C-Mixer}$ takes an input of the same size.
Furthermore, notably, we do not consider the role of positional codes.
2) Pre-training.
We follow the standard self-supervised learning procedure: pre-training followed by fine-tuning the downstream tasks.
Before fine-tuning $\texttt{C-Mixer}$, we require pre-training $\texttt{C-Mixer}$ on a training set without labels.
We pre-train the model at resolution 28 $\times$ 28 using AdamW with $\beta_{1}$ = 0.9, $\beta_{2}$ = 0.999, linear learning rate warmup of 1000 steps and linear decay, batch size 500.
For the random masking rate in self-supervised learning, we use 20\%, and the random seeds are not fixed.
For the target view, we pre-process images by applying the cropping technique, random horizontal flipping, and contrast conversion. 
The similarity metrics of the anchor view and target view using only the cross-entropy function.
3) Fine-tuning. We fine-tune using momentum SGD, batch size 512, gradient clipping at global norm 1, and cosine learning rate schedule with a linear warmup.
Note that the whole process of fine-tuning does not use the data augmentation scheme or early-stopping strategy.

\noindent \textbf{Ablation study.}
We conduct several ablation experiments to demonstrate the effectiveness of each module.
1) The effectiveness of self-supervised algorithms.
We perform two subtasks: without self-supervised learning (w/o ssl); without random masking (w/o rm).
For without self-supervised learning, we only train the network at resolution 28 $\times$ 28 using AdamW with $\beta_{1}$ = 0.9, $\beta_{2}$ = 0.999, linear learning rate warmup of 1000 steps and linear decay, batch size 500.
For without random masking, we only remove the random masking operation during the pre-training process.
2) The effectiveness of incentive learning (w/o il).
We use a standard MLP-Mixer (we use a ``B'' architecture) to replace $\texttt{C-Mixer}$. 
The imaginary component that is imposed on the learner is removed.
3) The effectiveness of Pearson curves (w/o \texttt{p}(r), w/o \texttt{p}(i)).
We conduct two ops to replace $\texttt{p}$, one to keep only the real part and the other to keep only the imaginary part.

\begin{figure}[h]\scriptsize
	\begin{center}
		\tabcolsep 1pt
		
		\vspace{-2mm}
		\begin{tabular}{@{}ccc@{}}
			\includegraphics[width = 0.16\textwidth]{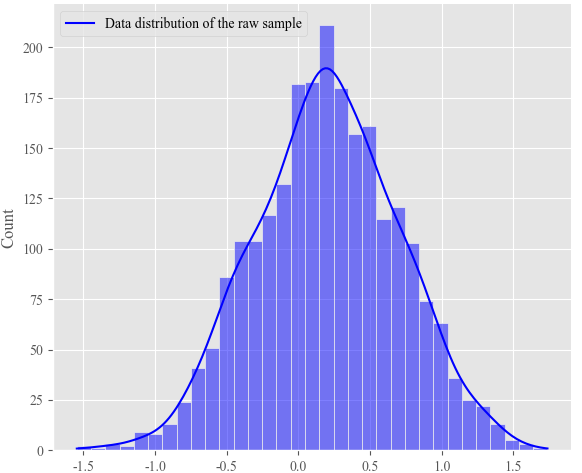}                &
			\includegraphics[width = 0.16\textwidth]{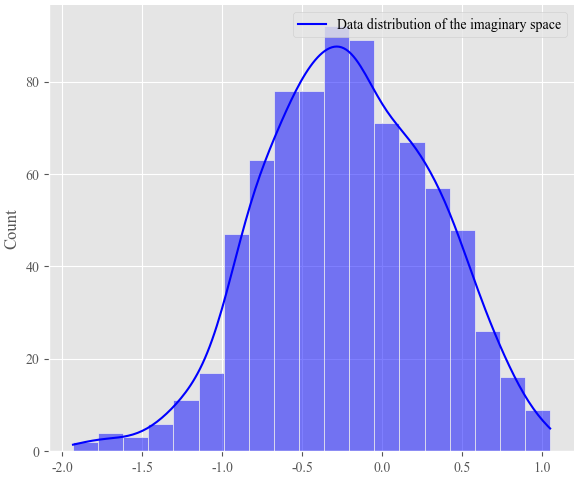}          &
			\includegraphics[width = 0.16\textwidth]{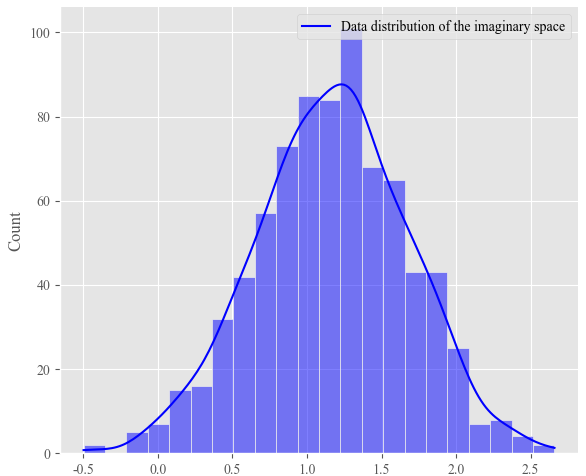}            \\
			
			\includegraphics[width = 0.16\textwidth]{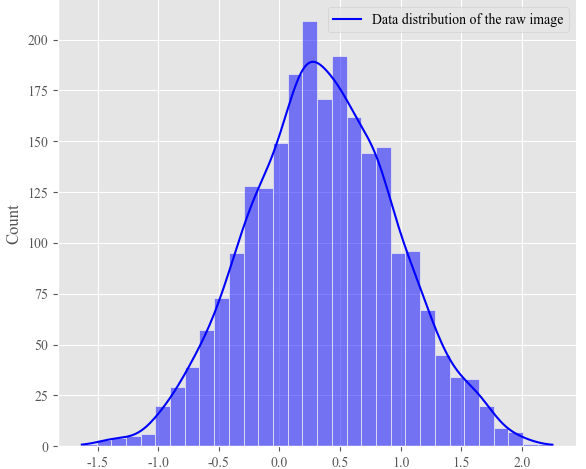}                &
			\includegraphics[width = 0.16\textwidth]{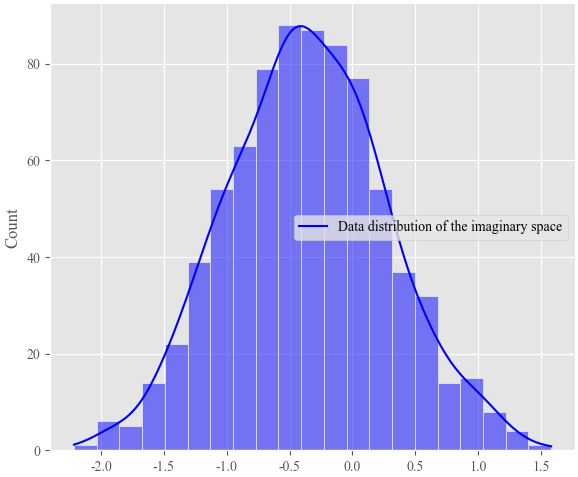}          &
			\includegraphics[width = 0.16\textwidth]{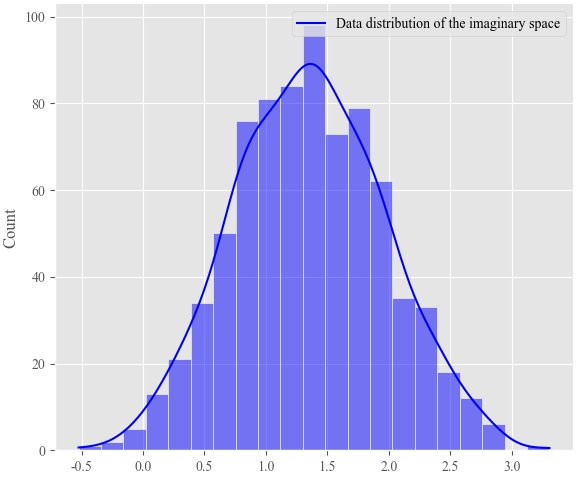}            \\
			
			(a)     &
			(b)     &
			(c)    \\  
		\end{tabular}
	\end{center}
	\vspace{-4mm}
	\caption{This figure shows the data distribution of a raw image, corresponding to a pair of imaginary distributions. Among them (b) indicates the generated imaginary distribution when the prediction label is 1, and (c) indicates the generated imaginary distribution when the prediction label is 0. \textbf{For the best visual representation, all numbers are normalized to [-3, 3].}}
	\vspace{-4mm}
	\label{figs}
\end{figure}

\noindent \textbf{Discussion and analysis.}
Our proposed method achieves competitive performance on most of the data sets (including three types of tasks), however, one question still needs to be discussed, i.e. \textit{what kind of data distribution is provided by the imaginary space ($\Omega$)}?
Several samples are drawn to demonstrate the data distribution by applying the visualization function.
As shown in Figure~\ref{figs}, we note that the distribution of imaginary numbers is symmetric concerning the distribution of real numbers (raw image), and the axis of symmetry is the label value.
%
%
In general, the imaginary signal complements the distribution of information (non-standard normal distribution) that is symmetrical or not provided by the raw image.

\vspace{-2mm}
\section{The potential of algorithms.}
\vspace{-2mm}
We conduct several experiments to demonstrate the potential of incentive learning and self-supervised learning.

\noindent \textbf{Incentive learning.} 
The role of incentive learning is primarily demonstrated to complement the components of the input information such that the learning cost of the learner is reduced.
We propose this learning paradigm which is very suitable for generating dense prediction tasks, such as exposure correction and dehazing, where we use U-Net~\cite{RonnebergerFB15} as the learner.
The final output of the network is still enforced with a generalized Pearson curve $\texttt{P}$ to project into the real number field.
The U-Net is implemented in PyTorch 1.2 and AdamW optimizer is used to train the network.
We use the resolution 512 $\times$ 512 images with batch size 8 to train the network, the learning rate is set to 0.001.
We only use L1 for the loss function that is required to train the network.
As shown in url~\url{https://www.researchgate.net/profile/Zhuoran-Zheng?ev=hdr_xprf}, we show some examples of image enhancement.
We observe that algorithms with imaginary complementary components are more advantageous in restoring color and texture, where the convolution operator of U-Net uses complex convolution.
For a fair comparison, U-Net without complex convolution adds a large number of convolution operators (the convolution kernel is a 3 $\times$ 3 with a step stride of 1) at the tail of the model to keep the computational power balanced.

\noindent \textbf{Self-supervised learning.} 
Our proposed strategy is compared with Masked Siamese Network (MSN)~\cite{AssranCMBBVJRB22} on several datasets.
MSN is a self-supervised learning algorithm with prototype learning, but it only works with a random masking scheme enforced in the anchor view.
We use a standard Vision Transformer (ViT)~\cite{dosovitskiy2020image} as the encoder.
For the loss function, we use cross-entropy.
Our proposed method with ViT vs. MSN with ViT is both pre-trained on the ImageNet dataset and then fine-tuned to train on the CIFAR10 and CIFAR100 datasets in an end-to-end manner.
Models transferred to CIFAR10 and CIFAR100 are fine-tuned for 1000 epochs using a batch size of 600 and a learning rate of 0.0001.
We use the data augmentations defined by RandAugment.
For CIFAR10 and CIFAR100, our proposed method boosts the performance of MSN by \textbf{2\%} (ACC) compared to MSN, however, the AUC is slightly lower.
The probable reason is that our method bears a higher degree of stochasticity (random masking) compared to MSN.
Besides, by applying our method and MSN on the MNIST dataset (fine-tuned in MNIST), the encoder is still a standard ViT.
Here the training dataset of MNIST is only 10\% of the original dataset, which contains samples for each class.
The experimental results show that our method is almost consistent with the MSN performance.

\section{Related Work}
\noindent \textbf{MedMNIST classification.}
Most of the medical image datasets are created with background knowledge from vision or medicine.
This makes it difficult for most algorithm engineers, and to avoid this problem, two versions of \textit{MedMNIST} (v2)~\cite{yang2023medmnist} are released one after another.
To sound this community, the developers leverage two models, ResNets~\cite{he2016deep} and AutoML~\cite{he2021automl}, as the classification algorithm for \textit{MedMNIST}.
Following that, FPVT~\cite{liu2022feature} and MedViT~\cite{manzari2023medvit} are proposed to establish a new baseline by modeling long ranges on medical images.
In addition, \textit{MedMNIST} is also used to evaluate the performance of the algorithm for federal learning tasks~\cite{lu2021distribution,wang2021heterogeneous}.
Here, we use \textit{MedMNIST} (v2) only for validating the learning ability of the network.

\noindent \textbf{Complex filed learning.}
Recently, approaches based on modeling complex domains are mentioned~\cite{choi2019phase,hu2020dccrn,quan2021image,trabelsi2017deep}.
These methods mainly involve representation by performing domain transformation on the input signal with the help of deep learning.
To boost the performance of the classifier, we consider modeling over the complex domain.

\noindent \textbf{Mixer network.}
MLP-Mixer~\cite{TolstikhinHKBZU21} proposes a conceptually and technically simple architecture solely based on MLP layers. 
However, the standard MLP-Mixer performs weakly on medium-scale datasets (including \textit{ImageNet-1K} and \textit{ImageNet-21K}).
%
To further boost the performance, a pure MLP architecture is proposed for this network, called Res-MLP~\cite{touvron2022resmlp}. 
gMLP~\cite{zhang2021gmlp} designs a gating operation to enhance the communications between spatial locations and achieves a comparable recognition accuracy compared with Res-MLP. 
More advanced works~\cite{tang2023mlp,yu2022s2} adapt a hierarchical pyramid to enhance the representing power.

\noindent \textbf{Self-supervised learning.}
The goal of self-supervised learning is to allow the learner to still generalize well for training sets that contain little or noisy labeling.
The typical methods are~\cite{baevski2022data2vec,misra2020self}, and \cite{elnaggar2021prottrans}, they aim to mine the common semantics of the input information.
The current state-of-the-art model is MSN~\cite{AssranCMBBVJRB22}, which introduces mask learning.

\section{Conclusion}
In this paper, we alleviate two inherent drawbacks of \textit{MedNMIST} (v2) effectively via the designed incentive learning and self-supervised learning methods with random masking. 
In addition, we use incentive learning to evaluate the image enhancement task with significant gains and self-supervised learning with random masking with great potential.
Extensive experimental results show that our method performs well on supervised tasks, weakly supervised tasks, and dense prediction tasks.
In future work, we apply this new machine learning paradigm to more generative or inferential tasks.

\bibliography{refs.bib}
\bibliographystyle{ieee_fullname.bst}
\vfill\pagebreak

\end{document}